\documentclass[final]{cvpr}

\usepackage{times}
\usepackage{epsfig}
\usepackage{graphicx}
\usepackage{amsmath}
\usepackage{amssymb}
\usepackage{bbm}
\usepackage{color}
\usepackage{multirow}
\usepackage{subcaption}
\usepackage{pifont}
\newcommand{\cmark}{\ding{51}}%
\newcommand{\xmark}{\ding{55}}%

\newcommand{\Tref}[1]{Table~\ref{#1}}

\newcommand{\Fref}[1]{Figure~\ref{#1}}

\usepackage[pagebackref=true,breaklinks=true,colorlinks,bookmarks=false]{hyperref}

\def\cvprPaperID{669} 
\def\confYear{CVPR 2021}

\begin{document}

\title{Identity-Driven DeepFake Detection
}

\author{Xiaoyi Dong$^{1}$\thanks{Work done during an internship at Microsoft Research Asia.},  Jianmin Bao$^{2}$, Dongdong Chen$^{2}$, Weiming Zhang$^{1}$,\\ Nenghai Yu$^{1}$, Dong Chen$^{2}$, Fang Wen$^{2}$, Baining Guo$^{2}$ \\

$^{1}$University of Science and Technology of China 
$^{2}$Microsoft Research\\

{\tt\small\{dlight@mail., zhangwm@, ynh@\}.ustc.edu.cn } 
{\tt\small cddlyf@gmail.com }\\
{\tt\small\{jianbao, luyuan, doch, fangwen, bainguo \}@microsoft.com } 
}

\maketitle

\begin{abstract}
DeepFake detection has so far been dominated by ``artifact-driven'' methods and the detection performance significantly degrades when either the type of image artifacts is unknown or the artifacts are simply too hard to find. 
In this work, we present an alternative approach: Identity-Driven DeepFake Detection. Our approach takes as input the suspect image/video as well as the target identity information (a reference image or video). We output a decision on whether the identity in the suspect image/video is the same as the target identity. Our motivation is to prevent the most common and harmful DeepFakes that spread false information of a targeted person. The identity-based approach is fundamentally different in that it does not attempt to detect image artifacts. Instead, it focuses on whether the identity in the suspect image/video is true.  To facilitate research on identity-based detection, we present a new large scale dataset ``Vox-DeepFake", 
in which each suspect content is associated with multiple reference images collected from videos of a target identity.
We also present a simple identity-based detection algorithm called the OuterFace, which may serve as a baseline for further research. Even trained without fake videos, the OuterFace algorithm achieves superior detection accuracy and generalizes well to different DeepFake methods, and is robust with respect to video degradation techniques -- a performance not achievable with existing detection algorithms.

\end{abstract}

\section{Introduction}

As a new face forgery technique, DeepFake can replace the face of a source image with a target face in a highly convincing way and thus generate fake videos of actions never performed by the targeted individual. The malicious usage and spread of DeepFake~\cite{DeepFaceLab, DFaker,FakeApp,faceswap, faceswapGAN,li2019faceshifter,bao2018openset,nirkin2019fsgan,nirkin2018face} have raised serious societal concerns, especially when the faked identity is a politician or celebrity. Therefore, how to detect such malicious DeepFake content has become a hot research topic recently.  

\begin{figure}[t]
	\centering
	\includegraphics[width=1\columnwidth]{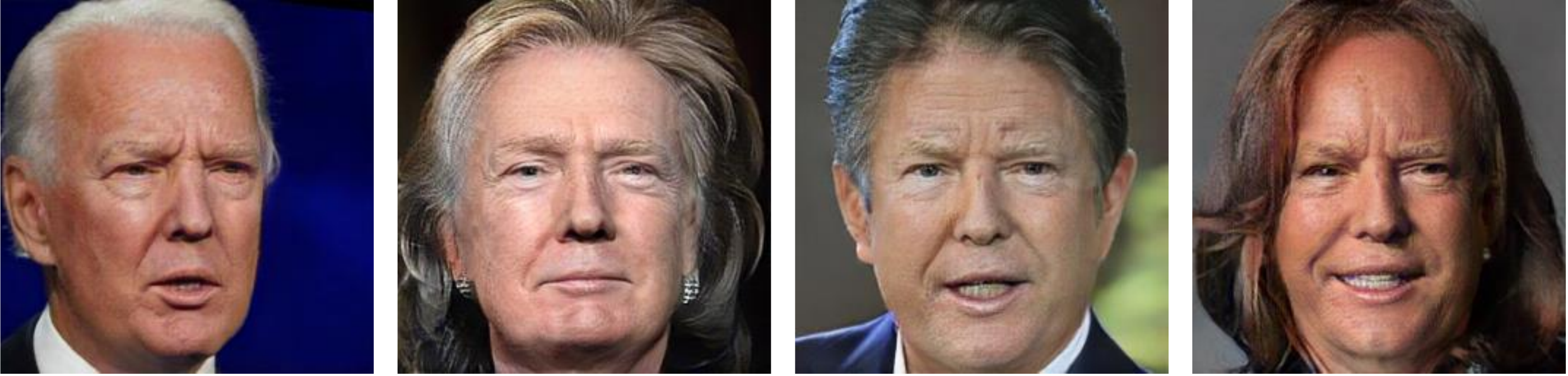} 
	\caption{Finding image artifacts in DeepFakes is not always easy. Existing methods fail to detect artifacts in the above DeepFake images. With our identity-driven approach, we can easily detect these as fake.}
	\label{fig:intro}
\end{figure}

To date, most existing detection methods are ``artifact-driven": they try to discriminate fake images by searching for underlying generation artifacts \cite{zhou2018twostream,zhou2018learning,liu2018image,bappy2019hybrid,faceforensics,li2019exposing,afchar2018mesonet,matern2019exploiting,nguyen2019multitask,nguyen2019use,li2020face,qian2020thinking} or unnaturalness of DeepFake generation methods \cite{yang2018exposing,li2018ictu}. While these methods perform well for the close-set setting, the performance degrades significantly when the type of artifacts is unknown (open-set) or artifacts are too hard to find as the generation quality continues to improve. Take the four carefully crafted images in \Fref{fig:intro} as examples, we tried several state-of-the-art detection methods \cite{li2019exposing,afchar2018mesonet,faceforensics} and found they all failed.

We observe that a critically important factor of DeepFake detection is the identity of the person in DeepFake images and videos, and the most common and harmful DeepFakes are those which spread false information of a targeted individual, especially if the individual is a politician or celebrity. Based on this observation, we propose a new approach by explicitly including identity as part of the DeekFake detection. With the identity serving as a strong semantic-level prior, we aim to make DeepFake detection an easier problem. It is interesting to note that common people are already using identity in their efforts to spot DeepFakes, albeit naively, by taking advantage of the attribute prior knowledge associated with the identity, \ie, the usual hairstyles of a celebrity.

Our new approach, called Identity-Driven DeepFake Detection, is formulated as follows. Given the suspect image/video as well as the target identity information (a reference image or video), we compute whether the identity in the suspect image/video is the same as the target identity. Essentially instead of the existing \textit{``artifact-driven"} detection methodology, we reformulate a \textit{``identity-driven"} one. Rather than trying to find generation artifacts, we model DeepFake detection as an identity verification problem, \ie, \textit{``whether the suspect image has the same identity as the reference image"}. In practice, the identity information can be obtained through the surrounding context (e.g., attached text in social media), face recognition models, or a reference image provided by a trusted regulatory platform.

This identity-driven approach shares the spirit of the popular face verification technique. However, deep down there is a big difference because most DeepFake methods can generate high-quality inner face parts, and existing face verification methods \cite{deng2019arcface,wang2018cosface}, which also focus on these parts, do not perform well in discriminating real/fake faces, as we empirically verified. It will be even harder for face verification methods to succeed in the future as the generation quality of DeepFake continues to improve.

To facilitate research on identity-driven DeepFake detection, we have built a new large scale DeepFake detection dataset ``Vox-DeepFake", which has a total of 2M real videos and fake videos. Compared to existing datasets, ours is not only the first that supplies the explicit reference identity information for each real/fake video, but also has better quality and diversity in terms of identities and video contents.  
To simulate different generation methods, the fake videos are generated with three different state-of-the-art methods DeepFake~\cite{faceswap}, FSGAN~\cite{nirkin2019fsgan}, and FaceShifter~\cite{li2019faceshifter}. For simplicity, we directly provide a real image with the same identity as the reference prior in the dataset.

We further propose a new identity-driven algorithm ``OuterFace", which may serve as a strong baseline for further research. The algorithm deliberately erases parts of the inner face regions by using the landmark guidance and thereby forces the train model to learn a robust representation of the outer face region for verification. While the algorithm is simple and can certainly be improved in many ways, its experimental results are quite strong. Even without the need of training on any fake video, it achieves superior detection accuracy and generalizes well to different DeepFake generation methods. More surprisingly, it is also robust to noise and video compression codec mechanism. The model only trained on the VggFace2 dataset~\cite{cao2018vggface2} and can achieve more than $96\%$ AUC on three previous datasets. A similar high performance is achieved when the model is directly applied to highly compressed videos. These are strong results that are unachievable with existing ``artifact-driven" methods, and we believe that with more finetuning with fake videos, the performance of our model can be further improved.

\section{Related Work}
\noindent \textbf{DeepFake Generation}.
DeepFake generation is the general term for face swapping algorithms. The face swapping algorithms are roughly divided in three categories: landmark-based~\cite{bitouk2008face,wang2008facial}, 3D-based~\cite{blanz2004exchanging,wang2008facial,cheng20093d,dale2011video,lin2012face,nirkin2018face,nirkin2019fsgan} and GAN-based~\cite{korshunova2017fast,faceswap,natsume2018fsnet,natsume2018rsgan,bao2018openset,olszewski2017realistic,li2019faceshifter} approaches. Early efforts~\cite{bitouk2008face,wang2008facial} are landmark-based which only swap faces with similar poses. To address such limitation, many follow-up works~\cite{blanz2004exchanging,wang2008facial,cheng20093d,lin2012face} consider 3D transform between two faces. To preserve the target facial occlusions, Nirkin \etal~\cite{nirkin2018face,nirkin2019fsgan} train an occlusion-aware face segmentation network to predict visible target face mask for blending. Recently GAN-based methods achieve more vivid face swapping results. Korshunava \etal~\cite{korshunova2017fast} and DeepFakes~\cite{faceswap} train a transformation model for each target person. RSGAN~\cite{natsume2018rsgan}, IPGAN~\cite{bao2018openset} and FaceShifter~\cite{li2019faceshifter} addressed this limitation by subject-agnostic face swapping researches. Although the quality of fake images is constantly improving, all DeepFake generation methods can only generate the inner face area. The outer face still comes from the target image. This makes our outer face identity-driven detection particularly effective.

\noindent \textbf{DeepFake Detection}.\label{sec:detection}
In the past two years, many different DeepFake detection methods \cite{zhou2018twostream,zhou2018learning,liu2018image,bappy2019hybrid,faceforensics,li2019exposing,afchar2018mesonet,matern2019exploiting,nguyen2019multitask,nguyen2019use,li2020face,qian2020thinking,yang2018exposing,li2018ictu} have been proposed. Fundamentally, they can be all regarded as artifact-driven. They either aim to find the explicit appearance artifacts like the texture difference and blending ghost \cite{bappy2019hybrid,faceforensics,li2019exposing,li2020face}, or identify the physical unnaturalness like the head pose difference between inner and outer face \cite{yang2018exposing} and the eye blinking error \cite{li2018ictu}. Despite the good performance in the close-set setting, they suffer from a serious performance drop when transferring to unseen generation methods, datasets, and image/video degradation. More importantly, as fewer artifacts appear in future generation methods, such artifact-driven methods will continue to struggle.  By contrast, we observe identity information is a very useful and essential prior, and propose an identity-driven approach, which demonstrates better accuracy and much stronger transferability

\begin{table*}[t]
\begin{center}

\setlength{\tabcolsep}{1.9mm}{
\begin{tabular}{c|c|c|c|c|c|c|c}
\hline
Dataset  & \#Fake Video & \#Real Video & \#Id & \#Unique Video/Id & \#Method & Explicit Id Ref & Avg AUC \\
\hline\hline
UADFV               & 49        & 49      & 1        & 1    & 1   & \xmark & 78.51\\
DF-TIMIT            & 640       & 320     & 43       & 10   & 2   & \xmark & 75.52\\
FaceForensics++     & 4000      & 1000    & 1000     & 1    & 2   & \xmark & 86.56\\
Google DFD          & 3000      & 363     & 28       & 12.9 & 1   & \xmark & 79.23\\
Celeb-DF            & 5639      & 590     & 59       & 10   & 1   & \xmark & 66.75\\
DeeperForensics-1.0 & 1000      & 1000    & 1000     & 1    & 1   & \xmark &  62.58\\
DFDC                & 104,500   & 23,654   & 960      & 1    & 4  & \xmark &  65.69 \\
\hline
Vox-DeepFake        & \textbf{1,045,786} & \textbf{1,125,429} & \textbf{4000}     & \textbf{26.2} & 3   & \cmark & \textbf{56.31}\\
\hline
\end{tabular}}
\end{center}
\vspace{-0.5cm}
\caption{The statistics comparison between our Vox-DeepFake and existing DeepFake datasets. `Explicit Id Ref' means whether the identity between real and fake videos are provided explicitly. The `Unique Video' here means the videos are captured from totally different scenes, instead of different splits of a long video. `Avg AUC' is the mean detection AUC calculated by 9 open-source methods.}
\label{tbl:dataset_list}
\end{table*}

\noindent \textbf{DeepFake Datasets}.\label{dataset_rel}
To speed up the development of DeepFake detection,  a lot of benchmark datasets appear, such as UVDFV~\cite{korshunov2018deepfakes}, FaceForenscics++ ~\cite{faceforensics} , DeepFake-TIMIT~\cite{yang2018exposing}, Celeb-DeepFake~\cite{li2020celebdf}, Google DeepFake detection dataset DFD~\cite{google}, and FaceBook DeepFake detection challenge dataset DFDC~\cite{dolhansky2020deepfake}.
However, none of them are suitable for our identity-driven approach because of their limited diversity on identity number, generation method, and unique real video number.
For example, UVDFV, DeepFake-TIMIT, Celeb-DF, DFD only use one DeepFake generation method. FaceForensics++ and DFDC only have one real video for each identity. By comparison, our Vox-DeepFake use three different state-of-the-art DeepFake generation methods and 4000 identities. And on average, each identity has 26.2 unique videos. Such great diversity not only well satisfies our setting but also benefits existing detection methods.

\section{Vox-DeepFake Dataset}
\label{sec:dataset}


To support our identity-driven approach, we build a new DeepFake detection dataset Vox-DeepFake for the evaluation and possible training needs. Our Vox-DeepFake dataset is built on top of VoxCeleb~\cite{nagrani2020voxceleb} dataset. 
Different from existing datasets, we provide an explicit identity relation between real and fake videos and have multiple unique real videos for diverse reference selection. 
When it comes to previous datasets, none of them are suitable for our identity-driven approach.
If we adopt the latest dataset DFDC, it does not provide the identity relation between the fake and real videos, so it is hard to apply our identity-driven approach to it.
If we adopt existing datasets like UADFV, FaceForensics++, or DeeperForensics-1.0 which has an implicit identity relation that we could collect it by extra processing, since each identity only has one unique video, the reference image is only possible from the same video as the training/evaluation clips. This will not only make the network easily collapse into a trivial solution (e.g., background comparison) but also cause information leak for evaluation.

\begin{figure}[t]
\centering
\includegraphics[width=1\columnwidth]{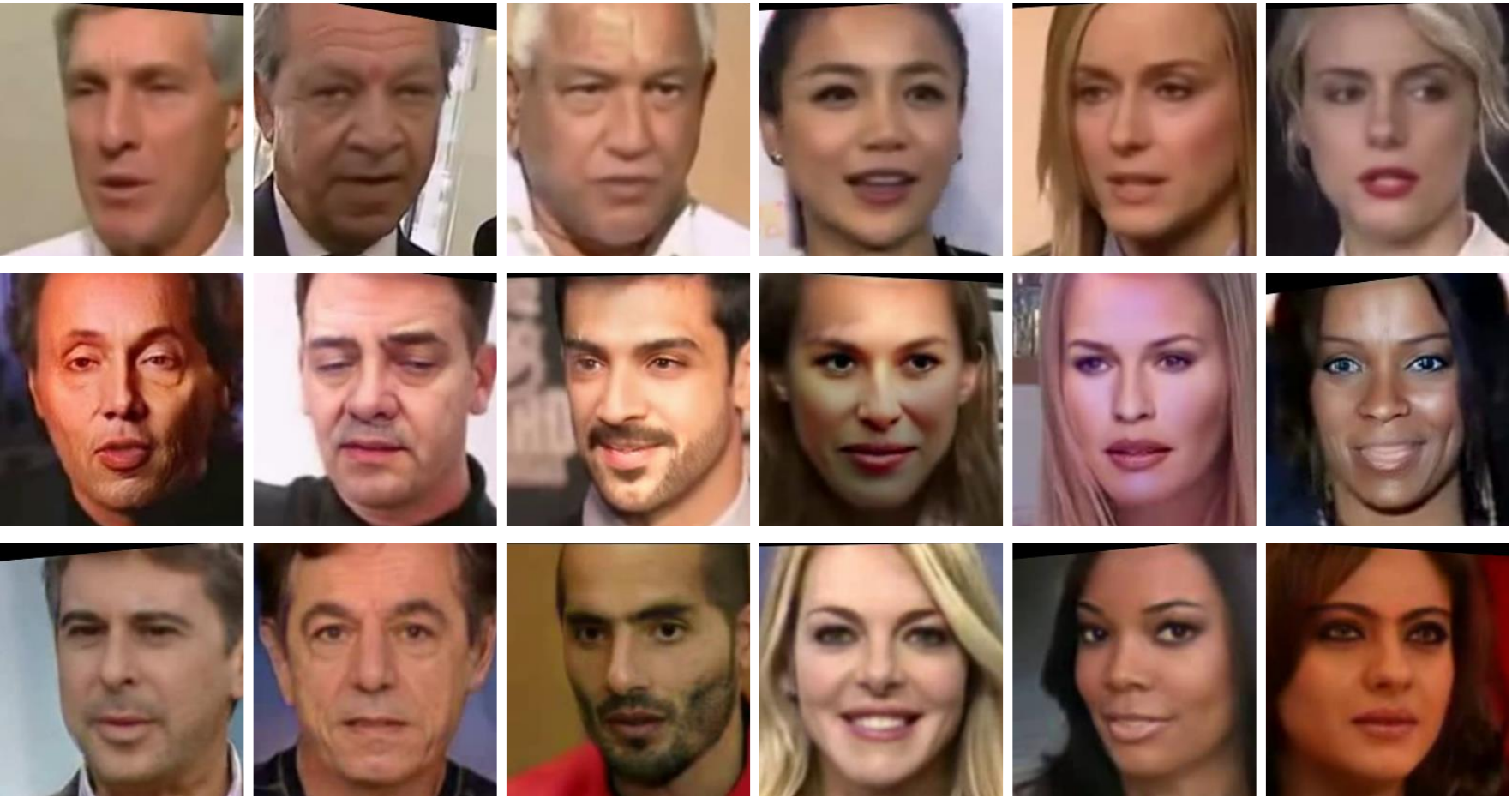}
\vspace{-6mm}
\caption{Example frames from our Vox-DeepFake dataset. The top to bottom rows are generated by DeepFake, FSGAN, FaceShifter respectively.}
\label{fig:vis_examples}
\end{figure}

Besides, in order to evaluate the cross-method performance, we use three different high-quality DeepFake generation methods to generate our fake videos, including DeepFake~\cite{faceswap}, FSGAN~\cite{nirkin2019fsgan}, and FaceShifter~\cite{li2019faceshifter}. While the existing dataset UADFV, DFD, Celeb-DF, and Deeperforences-1.0 only use one generation method. For DeepFake, it is the most popular generation method and swaps two faces with a shared auto-encoder structure. We use the implementation from the faceswap GitHub~\cite{faceswap}.
FSGAN and FaceShifter are the latest one-shot GAN-based method and their face swapping quality is much better than DeepFake, and we directly use their pretrained models. Some fake images generated by these three methods are shown in \Fref{fig:vis_examples}. 

\paragraph{Quantitative Comparison.} In \Tref{tbl:dataset_list}, we list the detailed statistics of existing datasets and our Vox-DeepFake. We follow the setting in DFDC~\cite{dolhansky2020deepfake} that clip all the videos into 10 seconds parts and report the number of the clipped videos. Obviously, our Vox-DeepFake has a larger scale and better diversity. In detail, it has 4000 identities and more than 1M fake videos, which are $4\times$ and $10\times$ larger than existing datasets. On average, there are 25.2 real unique videos for each identity, thus providing a greater diversity of the reference. We also report the average AUC calculated by 8 open-source DeepFake detection methods and our Vox-DeepFake is only 56.31, better than previous datasets. (please refer to the supplementary material for the detail).

\section{Identity-driven DeepFake Detection}
\label{sec:method}
With the identity-driven approach, we assume the identity of the suspect image/video is known in advance and provide an extra reference image as a prior. With this identity prior, we reformulate DeepFake detection as a \textit{``identity-driven"} problem, \ie, \textit{``does the suspect image have the same identity as the reference image?''}. This formulation is fundamentally different from existing DeepFake detection methods, which may be regarded as ``artifact-driven” since they always try to classify a suspect image as ``real/fake" by finding the underlying appearance artifacts or physical unnaturalness.

\begin{figure}[t]
\centering
\includegraphics[width=1\columnwidth]{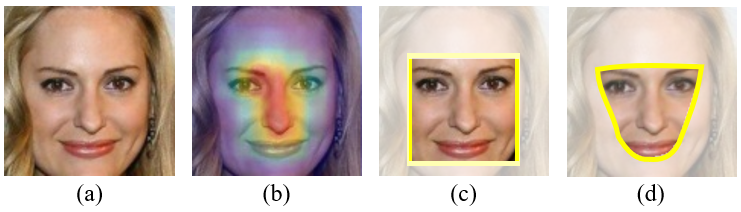} 
\vspace{-6mm}
\caption{(a) and (b) are the input and attentional region of the vanilla face verification model. (c) and (d) are the target region that existing DeepFake generation methods focus on to generate and replace.}
\label{fg:veri_visualization}
\end{figure}

A naïve way to perform identity-driven detection is to use traditional face verification techniques, which have been studied for decades. Unfortunately, our experiments show that face verification techniques are suboptimal for identity-driven DeepFake detection. On the Vox-DeepFake dataset, the AUC is only $83.41\%$. An important reason for this unsatisfactory performance is that traditional face verification heavily focuses on the inner face region. In \Fref{fg:veri_visualization} (b) we visualize the attention region where traditional face verification techniques focus on by using an occlusion-based saliency map. It can be seen that the model learned by traditional face verification regards the inner face region as the most discriminating part. This is a poor choice because existing DeepFake generation methods also focus on inner face generation and replacement as shown in \Fref{fg:veri_visualization} (c) and (d) and therefore tend to generate good quality images with few artifacts within this region. In particular, for face swapping methods the inner face region is a replacement directly from a real image and traditional face verification will fail to detect any problem in this region.

\begin{figure}[t]
\centering
\includegraphics[width=1\columnwidth]{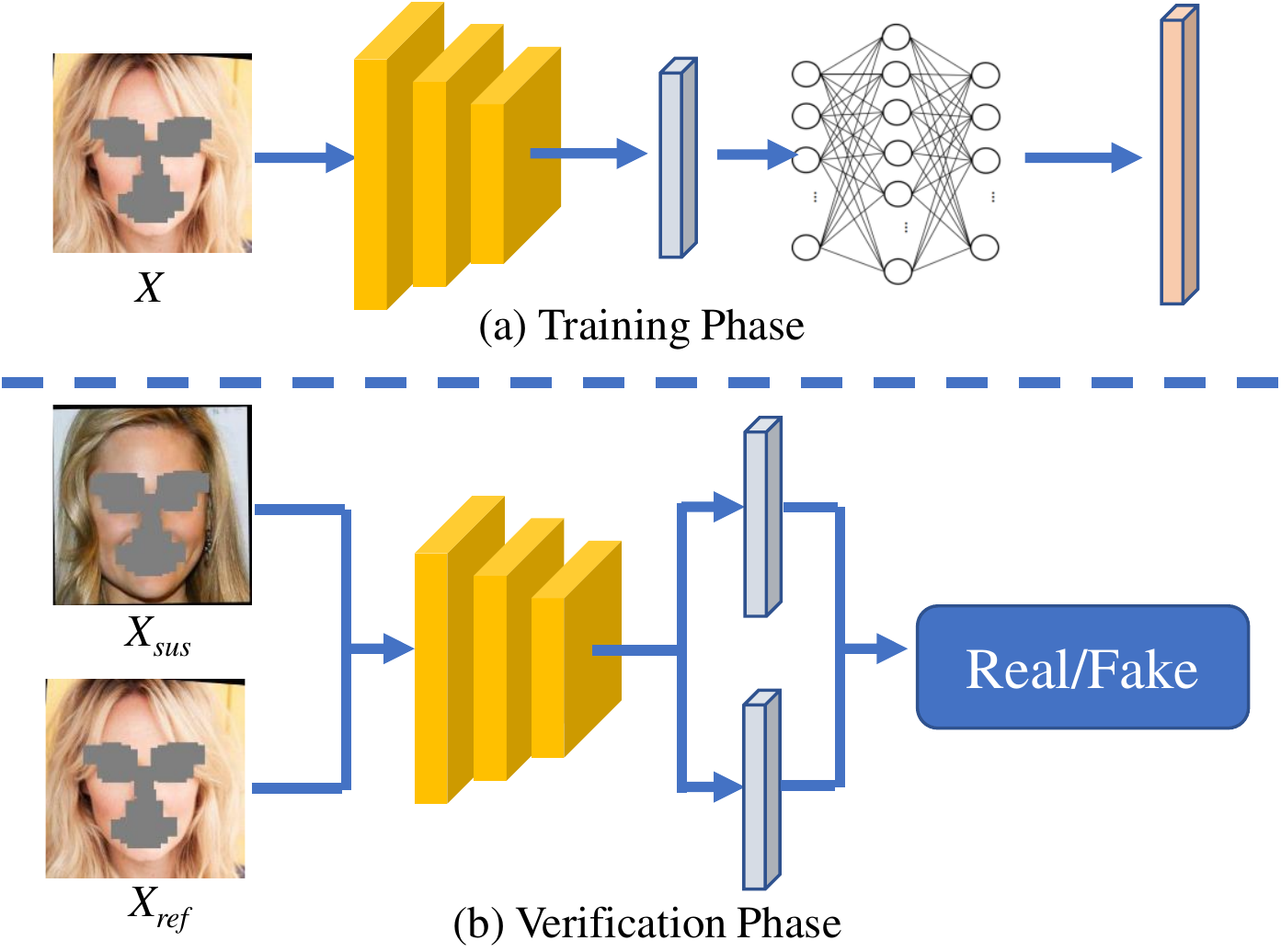}
\vspace{-6mm}
\caption{Pipeline of our OuterFace: (a) the identity embedding model is trained as a face recognition model during training, (b) the identity embedding distance between the suspect and prior reference image will be utilized for verification.}
\label{fig:pipeline}
\end{figure}

The identity-driven DeepFake detection framework provides a wide range of possibilities for designing highly effective detection algorithms. As an example we present \textbf{OuterFace}, a simple and effective algorithm and we hope this algorithm provides a solid baseline for further research. The core idea of the \textbf{OuterFace} algorithm is to learn a robust identity embedding from an expanded face region, which includes part of the inner face region but more importantly
the outer face region (the part of the whole head region beyond the inner face region). This design is based on two insights: 1) replacing the whole head properly is still difficult for existing generation methods, 2) the outer face region is very discriminating and should be included for identity verification. The overall pipeline is shown in \Fref{fig:pipeline}, which consists of two modules: landmark guided masking module for preprocessing and identity embedding model for verification.

\begin{figure}[t]
\centering
\vspace{-3mm}
\includegraphics[width=1\columnwidth]{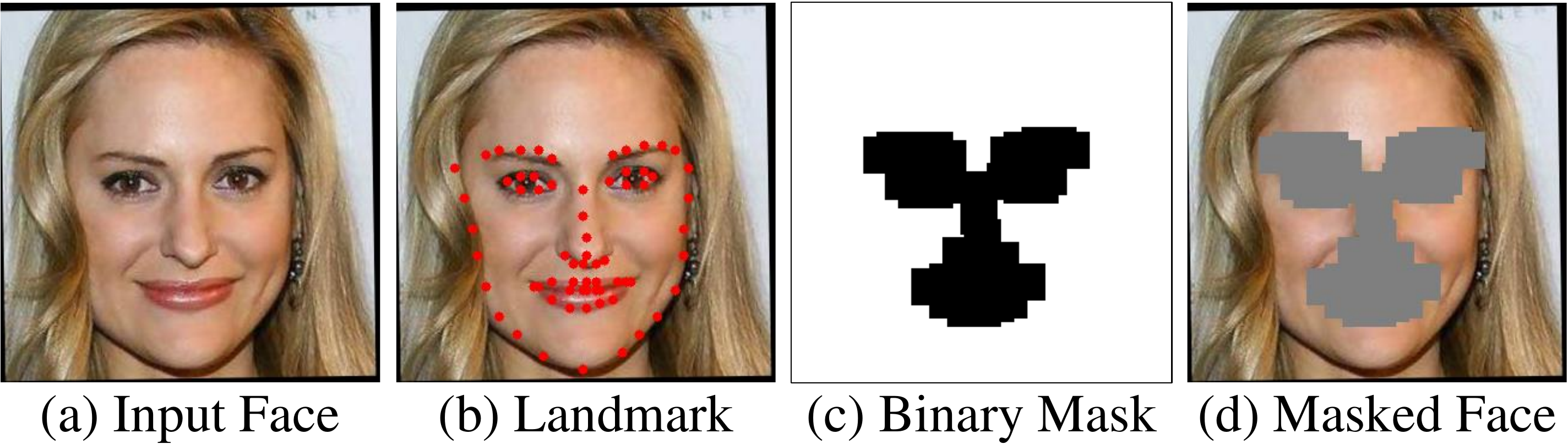} 
\vspace{-6mm}
\caption{Illustration of landmark guided masking: (a) input face, (b) the detected 68 landmarks, (c) landmark guided binary mask, (d) masked outer face.}
\label{fig:mask}
\end{figure}

\noindent\textbf{Landmark guided masking.} To make the verification model concentrate on the outer face region, we deliberately eliminate parts of the inner facial regions in both training and verification phases. Specifically, given a face image, we first use a face landmark detection model to estimate its 68-landmarks. Then we generate a binary mask with a selected subset of the landmarks. We mask out the neighbor pixels of each landmark point in the selected subset within a radius $k$. In our default setting, we use the point-wise mask with 51 landmark points, which is illustrated in \Fref{fig:mask}. Please refer to the supplementary material for the detailed selected landmark set.

\noindent\textbf{Identity Embedding Model.}
Similar to general face verification, we simply train the identity embedding model as a recognition model without bells and whistles. In detail, given a face image, we first eliminate the inner face out with the above landmark guided masking module, then feed it into the embedding model to extract the identity feature. To help training, an extra auxiliary projection head is added to project the embedding feature into the probability distribution among different identities. In our implementation, MobileNet~\cite{howard2017mobilenets} is adopted as the default embedding backbone and a fully-connected layer followed by softmax is used as the projection head. The input image size is $112\times 112$ and the feature embedding dimension is $512$.

Borrowing the knowledge from face recognition, we use the cosine-based softmax loss proposed in ArcFace \cite{deng2019arcface} rather than the vanilla cross-entropy loss for training.
\begin{equation}
    \mathcal{L} = -log \frac{e^{s \cdot cos(\theta_{i,j} + \mathbbm{1}\{j = y_i\} \cdot m)}}{\sum_{j=1}^{n}e^{s \cdot cos(\theta_{i,j} + \mathbbm{1}\{j = y_i\} \cdot m)}}
\end{equation}
Here $\theta_{i,j}$ is the angle between $W_j$ and $f_i$. $W_j \in \mathbbm{R}^d$ is the $j$-th column of weights of the final fully-connected layer in the projection head, $f_i \in \mathbbm{R}^d$ is the feature of $i$-th sample in the input batch. $\mathbbm{1}\{j = y_i\}$ is the indicator function that returns 1 when $j = y_i$ and 0 otherwise. $s$ is a scale hyperparameter and $m$ is an additive angular margin. By default, we set $s=64$ and $m=0.5$.

Similarly, during verification, to determine whether a suspect image has the same identity as the prior reference image, we mask the inner face out and extract their corresponding feature embeddings. Then the suspect image will be regarded as fake if their cosine distance is larger than a threshold $\tau$.

\paragraph{Fake-free Training.} In existing artifact-driven DeepFake detection methods, they all require a large scale of both real and fake data for training. It is because they need to learn the inherent artifact patterns. But by switching to identity-driven, our \textbf{OuterFace} algorithm shifts focus to identity verification and therefore does not need any fake content for training. In the following experiments, we only train our OuterFace on the face recognition dataset VggFace2~\cite{cao2018vggface2} by default, which has a total of 2.96 million images of 8361 identities. Thanks to this fake-free property, our OuterFace shows its strong generalization capability across different DeepFake generation methods and image/video degradation techniques.

\section{Experiment}
\label{sec:experimnet}

\noindent\textbf{Dataset setting.}
To demonstrate the strong generality of OuterFace, we do the evaluation on multiple different datasets: FaceForensics++ ~\cite{faceforensics} (FF++), Google DeepFake Detection~\cite{google} (DFD), Celeb-DeepFake~\cite{li2020celebdf} (Celeb-DF) and our Vox-DeepFake (Vox-DF).
For each dataset, we first crop out the face part with proper alignment for all the video frames. Then we select 20000 frames from the real and fake parts respectively. To confirm all the identities are covered, we select frames according to the identity evenly. For baseline artifact-driven methods, we evaluate their performance on the selected 40000 frames directly. 
For our identity-driven approach, we leverage the identity information and find a reference image for each identity. Since FF++ only has one real video for each identity, we have no choice but to choose the reference from the same source real video (excludes test frames). For all other datasets, we guarantee the reference images not to be selected from the same videos as test frames.

\noindent\textbf{Implementation detail.} 
In order to cover the outerface region, we crop a large face region for training. 
Specifically, we set the distance between two eyes as 27\% of the image width by default (Take \Fref{fig:mask} (a) as an example) and explore different ratios in the ablation study part.
The face classification model is trained from scratch at a resolution of $112 \times 112$ without any data argumentation. The total training epoch number is 30, and the batch size is 400. The initial learning rate is set to 0.1 and divided by 10 after 12, 15, 18 epochs. 

\begin{table*}[h]
\footnotesize
\begin{subtable}[h]{0.5\textwidth}
\centering
\begin{tabular}{l|c|cccc|c}

\hline
Method & Ori & C40 & JPEG &Resize & Noise & $\Delta$ \\
\hline\hline
Multi-task    & 72.23 & 62.00 & 61.52 & 55.90 & 63.21 & -11.57\\
Xception-c23*  & \textbf{98.54} & 87.25 & 86.87 & 93.53 & 77.01 & -12.37\\
DSP-FWA       & 81.90 & 61.97 & 59.99 & 79.23 & 65.64 & -17.18\\
\hline
CNNDetect     & 71.08 & 50.62 & 50.76 & 53.22 & 48.75 & -20.24\\
Patch-Foren*   & 73.75 & 73.53 & 65.91 & 56.27 & 65.98 & -8.327\\
Face X-ray*    & 98.44 & 72.76 & 80.61 & 79.23 & 71.58 & -22.39\\
\hline
InnerFace     & 98.22 & 95.89 & 98.21 & 98.29 & 98.18 & -0.577\\
OuterFace     & 98.24 & \textbf{97.97} & \textbf{98.23} & \textbf{98.34} & \textbf{98.24} & \textbf{-0.045}\\
\hline
\end{tabular}

\vspace{-0.12cm}
\caption{FaceForenscics++ ~\cite{faceforensics} }
\vspace{2mm}
\label{tab:ff}
\end{subtable}
\hfill
\begin{subtable}[h]{0.5\textwidth}
\centering
\begin{tabular}{l|c|cccc|c}
\hline
Method & Ori & C40 & JPEG &Resize & Noise & $\Delta$ \\
\hline\hline
Multi-task    & 65.21 & 49.44 & 47.65 & 48.01 & 46.26 & -17.37\\
Xception-c23 & 95.60 & 64.15 & 63.62 & 81.80 & 84.08 & -22.18 \\
DSP-FWA       & 90.99 & 58.78 & 74.62 & 65.73 & 66.07 & -24.69\\
\hline
CNNDetect     & 60.12 & 50.98 & 53.66 & 54.82 & 55.79 & -6.307 \\
Patch-Foren   & 49.91 & 60.01 & 50.13 & 50.10 & 50.91 & \textbf{2.877}\\
Face X-ray    & 94.14 & 54.99 & 64.87 & 58.36 & 63.78 & -33.64\\
\hline
InnerFace     & 97.82 & \textbf{96.88} & \textbf{97.59} & \textbf{97.89} & \textbf{98.05} & -0.217\\
OuterFace     & \textbf{97.93} & 96.87 & 97.20 & 97.51 & 97.82 & -0.580 \\
\hline
\end{tabular}
\vspace{-0.12cm}
\caption{Google DeepFake Detection~\cite{google}}
\vspace{2mm}
\label{tab:dfd}

\end{subtable}

\begin{subtable}[h]{0.5\textwidth}
\centering
\begin{tabular}{l|c|cccc|c}
\hline
Method & Ori & C40 & JPEG &Resize & Noise & $\Delta$ \\
\hline\hline
Multi-task    & 72.28 & 35.35 & 54.44 & 41.45 & 54.38 & -25.87 \\
Xception-c23$\;\;$    & 74.97 & 61.69 & 61.35 & 65.33 & 60.09 & -12.37 \\
DSP-FWA       & 78.51 & 51.48 & 58.33 & 66.04 & 57.05 & -20.28 \\
\hline
CNNDetect     & 56.12 & 39.11 & 55.80 & 54.13 & 49.33 & -6.527\\
Patch-Foren   & 59.66 & 60.12 & 67.95 & 60.63 & 52.49 & \textbf{0.637}\\
Face X-ray    & 74.76 & 55.52 & 67.36 & 60.90 & 62.26 & -13.25 \\
\hline
InnerFace     & 90.08 & 84.41 & 88.13 & 90.69 & 90.40 & -2.127\\
OuterFace     & \textbf{96.61} & \textbf{93.94} & \textbf{96.00} & \textbf{96.52} & \textbf{96.60}  & -0.692 \\
\hline
\end{tabular}
\vspace{-0.12cm}
\caption{Celeb-DeepFake~\cite{li2020celebdf}}

\label{tab:celebdf}
\end{subtable}
\hfill
\begin{subtable}[h]{0.5\textwidth}
\centering
\begin{tabular}{l|c|cccc|c}
\hline
Method & Ori & C40 & JPEG &Resize & Noise & $\Delta$ \\
\hline\hline
Multi-task    & 44.60 & 54.04 & 44.04 & 40.72 & 45.14 & \textbf{1.385} \\
Xception-c23  & 70.67 & 50.68 & 64.13 & 64.14 & 58.71 & -11.25\\
DSP-FWA       & 45.40 & 50.62 & 57.28 & 56.35 & 51.34 & 8.497 \\
\hline
CNNDetect     & 57.45 & 51.51 & 58.13 & 57.60 & 56.34 & -1.555 \\
Patch-Foren   & 51.82 & 55.14 & 54.96 & 53.96 & 47.85 & 1.157 \\
Face X-ray    & 65.08 & 51.35 & 77.21 & 55.74 & 60.19 & -3.957\\
\hline
InnerFace     & 83.41 & 82.17 & 83.76 & 83.65 & 83.41 & -0.162\\
OuterFace     & \textbf{90.61} &\textbf{ 88.27} & \textbf{90.31} & \textbf{90.60} & \textbf{90.12} & -0.785\\
\hline
\end{tabular}
\vspace{-0.12cm}
\caption{Vox-DeepFake}
\label{tab:vox}
\end{subtable}
\vspace{-2mm}
\caption{Frame-level AUC on different datasets and detection methods. Here `Ori' means images without any processing and C40, JPEG, Resize, Noise is different image degradation methods for robustness evaluation. $\Delta$ here means the average performance decrease caused by four strong image processing methods (Column 3-6 of each table). * indicates the model is trained and tested on the same dataset.}

\label{tab:main}
\end{table*}

\subsection{Comparison with Recent Works}

In this part, we make a comprehensive comparison with several representative works. Generally, these work can be divided into two groups, one group is specifically for the detection of images from the DeepFake method: Multi-task~\cite{nguyen2019multitask}, Xception-c23~\cite{faceforensics}, and DSP-FWA~\cite{li2019exposing}; the other group are three general DeepFake detection methods: Face X-ray~\cite{li2020face}, CNNDetection~\cite{wang2020cnn}, and Patch-Forensics~\cite{chai2020makes}. Meanwhile, we also report the performance of the traditional face verification method InnerFace, which is based on the idea of identity-driven and serves as the baseline of OuterFace.
As our method is a fake-free method, the dataset and training strategy is totally different from previous methods, so there is not a completely fair comparison on the dataset used in our method or previous methods. As a substitute, we focus on the generalization ability toward unseen datasets and image degradation for a relatively fair comparison. In fact, the generalization ability is not only the most important property of DeepFake detection for real-world usage but also the biggest challenge among existing methods. In the following experiments, we use the widely used AUC (area under the Receiver Operating Characteristic curve) as the evaluation metric.

\noindent\textbf{Generalization capability to unseen datasets.}
In real-world scenarios, it is common that the suspect images are generated by an unseen method or from an unseen dataset, so the generalization capability to unseen data is of great importance for DeepFake detection. In \Tref{tab:main}, we show the detection performance on four different datasets, where the close-set setting (\ie, train and test on the same dataset) is marked with ``*''. When comparing the results on the original images (Column ``Ori'' of each table), we find that our OuterFace generalize very well to different datasets and outperforms artifact-driven methods by large margins. Take Celeb-DF (\Tref{tab:celebdf}) as an example, the AUC of our OuterFace is 96.61\%, while the best AUC achieved by previous methods is only 78.51\%, nearly 18\% lower than us. Such great generalization ability indeed benefits from our identity-driven philosophy, which makes OuterFace not overfit to any specific DeepFake method.

\begin{figure}[t]
	\centering
	\includegraphics[width=1\columnwidth]{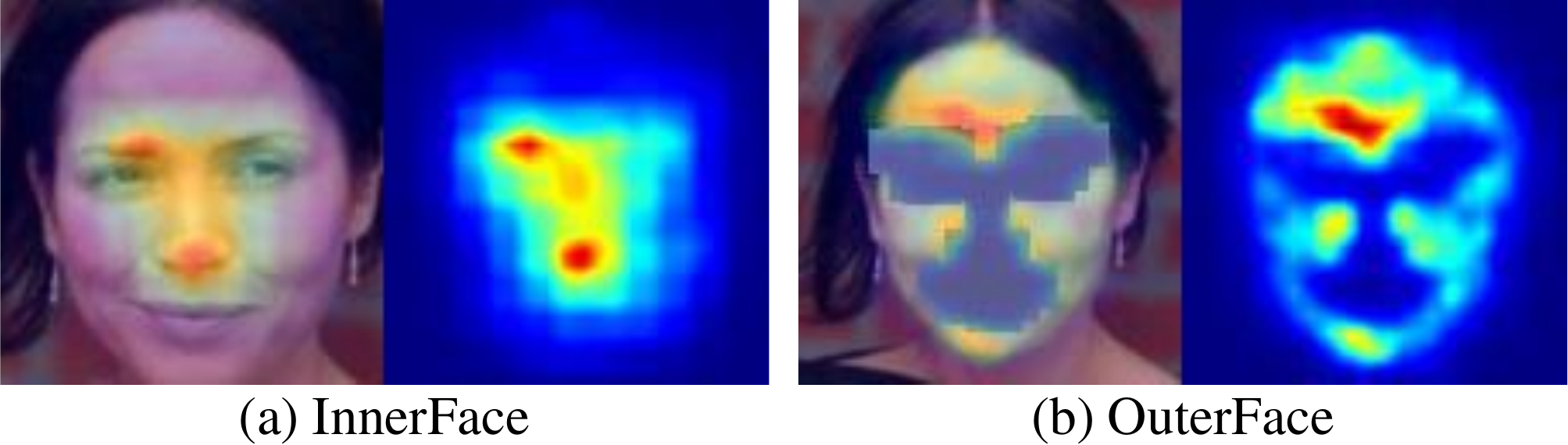} 
	\vspace{-6mm}
	\caption{Saliency map of InnerFace and OuterFace, area which has brighter color means the model pays more attention on it.}
	\label{fig:cam}
	\vspace{-5mm}
\end{figure}

\noindent\textbf{Generalization capability to image degradation.}
Image degradation is very common in the spread of DeepFake images and videos. In this part, we further evaluate the generalization capability to different kinds of image degradation. Here we consider both image-level degradation and video-level degradation, including: 1) JPEG-compression with quality factor 20; 2) Downsampling images $4\times$; 3) Gaussian noise with standard deviation 5; 4) Compressing the videos with high (C40) degrees of H.264 as FF++~\cite{faceforensics}. 

The results are reported in \Tref{tab:main} (Column 3-7 in each table, $\Delta$ is the average decrease compared to the results on original images). Obviously, with strong image degradation, the performance of artifacts-driven methods decreases significantly, while our OuterFace keeps robust with negligible performance drop (AUC decreases less than 1\% on average). Similar to the above cross-dataset robustness, it also benefits from our identity-driven setting. With the provided reference identity, our OuterFace focuses on the high-level semantic difference of the outer face part, which is invariant to low-level image processing.

\noindent\textbf{OuterFace \textit{v.s} InnerFace.} Under the proposed identity-driven framework, we provide more comparisons to the aforementioned identity-driven baseline ``InnerFace''. It works in a similar way as OuterFace but without the mask, so it uses the inner face region for detection. To make a fair comparison,  the same model architecture, loss function, and training set are used.

As shown in \Tref{tab:main}, we find that OuterFace performs better than InnerFace, and the performance gap increases with the increase of DeepFake generation quality. In detail, on earlier datasets such as FF++ and DFD, the performance of InnerFace and OuterFace is comparable, but on the higher-quality datasets Celeb-DF and Vox-DF, the performance gap becomes larger. 

To explore the reason, we use an occlusion-based saliency map to visualize what region the model focuses on. As shown in \Fref{fig:cam}, with a larger input area and our inner face mask, our OuterFace focuses on almost all the areas of the face, especially the forehead, which is unchanged during DeepFake generation. On the contrary, InnerFace mainly focuses on the most discriminative inner face part, which is carefully crafted by DeepFake, so its performance is related to generation quality.
We believe that as the quality of images generated by DeepFake improves, the gap between InnerFace and OuterFace will become larger and larger. 

From another perspective, InnerFace also has a very strong generalization ability, which further demonstrates the unique advantage of identity-driven DeepFake detection over existing artifact-driven counterpart.

\begin{table}[t]
\begin{center}
\small
\setlength{\tabcolsep}{1.9mm}{
\begin{tabular}{l|c|c|c|c|c}
\hline
Method &  Video1 & Video2 & Video3 & Video4 & Avg\\
\hline\hline
Multi-task   & 53.87 & 38.64 & 66.50 & 51.28 & 52.52 \\
Xception-c23 & 49.21 & 70.34 & 88.98 & 65.51 & 68.53\\
DSP-FWA      & 62.86 & 47.61 & 46.88 & 62.30 & 47.41\\
Face X-ray   & 54.67 & 84.21 & 86.87 & 48.82 & 68.64\\
\hline
OuterFace    & \textbf{100.00} &\textbf{ 99.72 }& \textbf{98.40} & \textbf{99.33} & \textbf{99.36}\\
			
\hline
\end{tabular}}
\end{center}
\vspace{-5mm}
\caption{DeepFake detection accuracy on videos collected from Internet. Here we report the mean accuracy on each video and the threshold used for real/fake prediction is concluded from the previous experiments.}
\vspace{-2mm}
\label{imitate}
\end{table}

\begin{table}[t]
\begin{center}
\small
\setlength{\tabcolsep}{1.85mm}{
\begin{tabular}{c|c|c|c|c|c}
\hline
Refer Strategy & FF++ &  DFD & Celeb-Df &Vox-DF  & Avg \\
\hline\hline
Random   & 98.24 & 97.93 & 96.61  & 90.61 & 95.85\\
Nearest  & \textbf{98.25} & \textbf{98.34} & \textbf{96.89}  & \textbf{96.71} & \textbf{97.53}\\
Farthest & 96.13 & 96.20 & 94.87 & 89.30 & 94.12\\
\hline
\end{tabular}}
\end{center}
\vspace{-5mm}
\caption{Analysis of different reference selection strategy. For Nearest and Farthest, we select the reference from 100 candidates based on the landmark Euclidean distance.}
\vspace{-2mm}
\label{pose}
\end{table}

\begin{figure}[t]
\centering
\includegraphics[width=1\columnwidth]{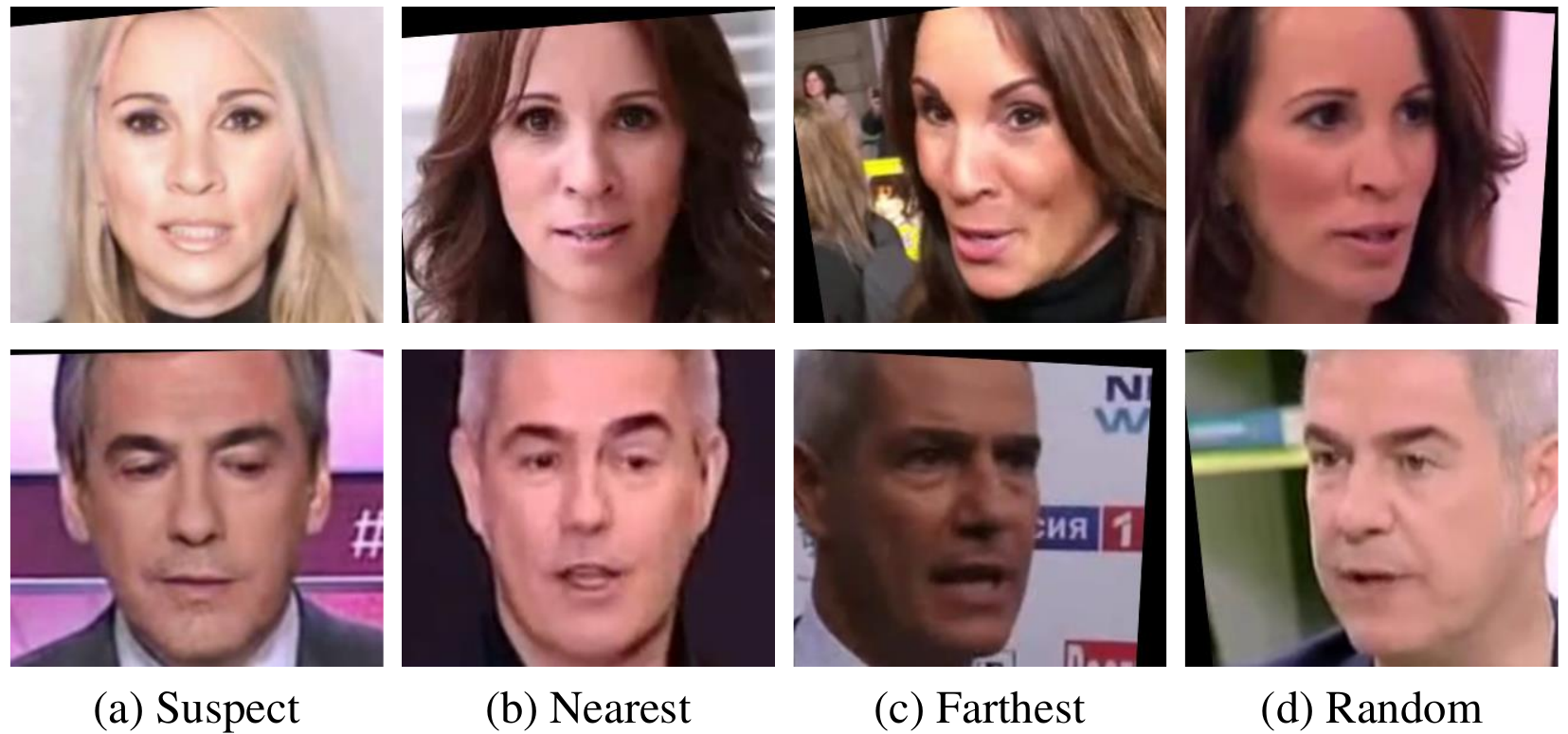} 
\vspace{-6mm}
\caption{Example of the reference image selected by different strategy. (a) The input suspect image. (b-d) The reference selected by Nearest, Farthest and Random strategy.}
\vspace{-3mm}
\label{fig:pose}
\end{figure}

\noindent\textbf{Real-world scenario evaluation}.
We further showcase one typical real-world DeepFake detection example by using our OuterFace: 1) We download four DeepFake videos from the YouTube channel ``Ctrl Shift Face\footnote{\url{https://www.youtube.com/channel/UCKpH0CKltc73e4wh0_pgL3g}}'', which are carefully crafted and all the faked identities are celebrities. 2) We split the videos into frames, then
use both the information from the video title and the online celebrity recognition model\footnote{\url{https://betaface.com/demo.html}} to get the faked identity information. 3) We search the reference images by Google with the identity name got in step 2. 4) We apply the reference image for our OuterFace and make the prediction. The threshold used in this part is concluded from previous experiments. 

With the results listed in \Tref{imitate}, we find that the accuracy of most existing methods is relatively low. But for our OuterFace, it is easy to predict whether the frame is real or fake and the accuracy is higher than 99\% on average. We also upload the video-version results on YouTube\footnote{\url{https://youtu.be/8nZfNbhuSPA}}.

\begin{figure}[t]
\centering
\includegraphics[width=1\columnwidth]{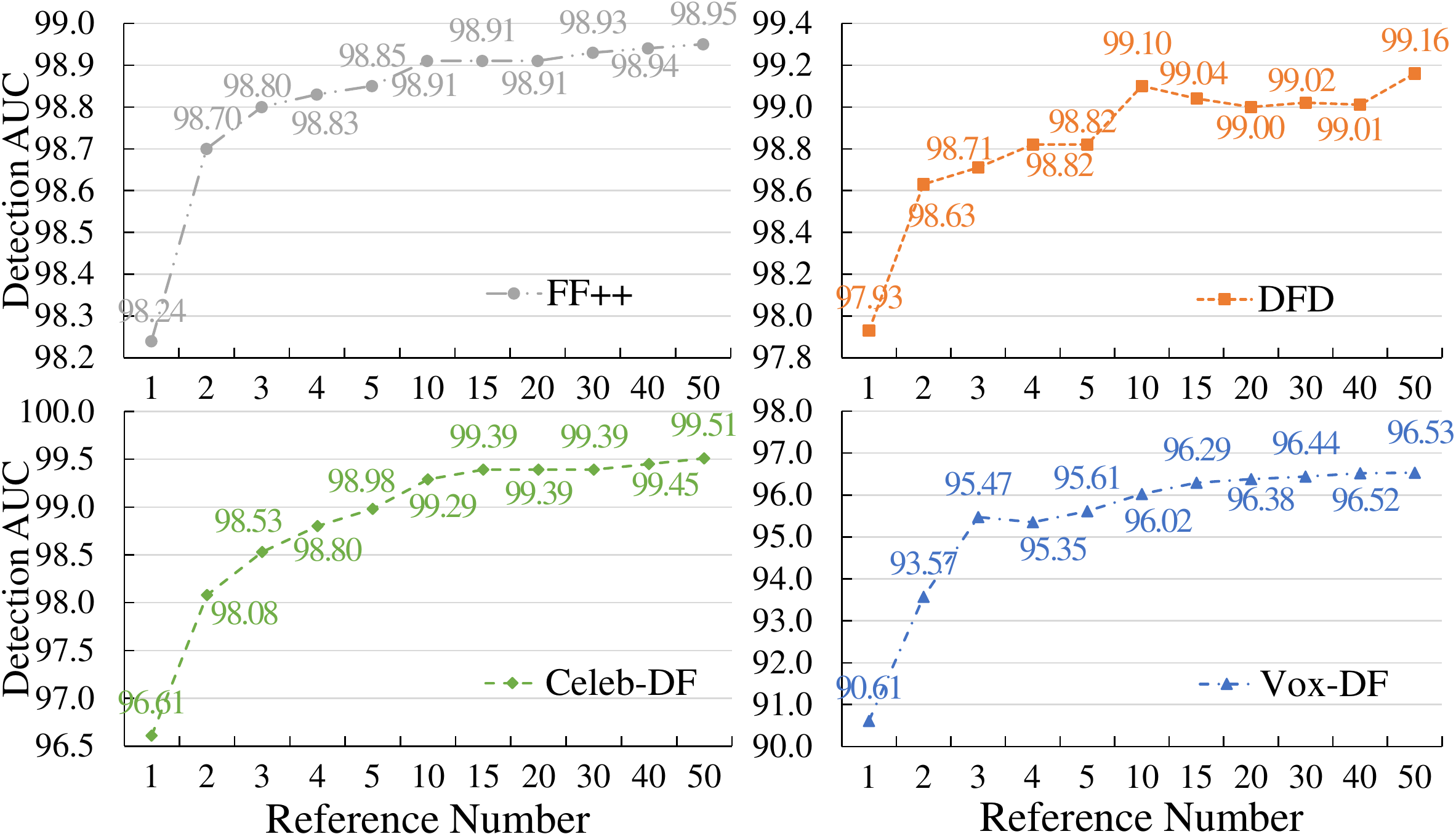} 
\vspace{-5mm}
\caption{The performance of OuterFace by using different number of reference images.}
\label{fig:refer}
\end{figure}

\subsection{Reference Sensitivity Analysis}
The core idea of our identity-driven DeepFake detection is the prior knowledge, more specifically, the reference image. So it is interesting to explore whether our OuterFace is sensitive to reference selection and whether the performance will increase by using more reference images.

\noindent\textbf{Sensitivity toward reference pose.}
As we all know, pose, expression, age, and lighting variations will affect the accuracy of face verification. These variations of the reference image may also influence the detection accuracy of OuterFace. In this part, we design some experiments to study the influence of the reference pose. In detail, for each suspect image, we randomly sample 100 reference images from different videos. Then we extract 68-landmarks for both suspect and reference images with a pre-trained face alignment model and use the Euclidean distance between landmarks as the pose similarity metrics. Finally, we select the reference image with three different strategies for testing: random, nearest, and farthest, as shown in \Fref{fig:pose}.

As shown in \Tref{pose}, without a doubt, the nearest pose setting achieves the best AUC on all the datasets. This is reasonable because when the suspect image and reference image have a similar pose, the identity embedding difference caused by pose difference is reduced. And the comparison could focus more on the identity difference and get a more accurate prediction. 
On the other hand, even by using the random or farthest reference, our OuterFace still works very well and the performance drop is only $2\%\sim~3\%$. Therefore, our OuterFace has a relatively low sensitivity to reference pose variation. Moreover, considering there also exist other variations in these three settings, such as expression and lighting variation, we believe our OuterFace is also robust to such reference variations.

\noindent\textbf{Sensitivity toward reference number.} In real applications, we may have many reference images for each target identity. Here we evaluate whether the performance of our OuterFace will become better if more than one reference image is provided.
In details, we vary the number of reference images from 1 to 50, and use the average identity embedding of all the reference images as the final identity embedding. As shown in \Fref{fig:refer}, as the number of references increases, the performance continues to improve. We guess the underlying reason may be similar to the above explanation of the performance fluctuation among pose variation, i.e., the ensemble of multiple identity embedding has reduced the influence from other face attributes.

\subsection{Ablation Study}

\begin{figure}[t]
	\centering
	\includegraphics[width=1\columnwidth]{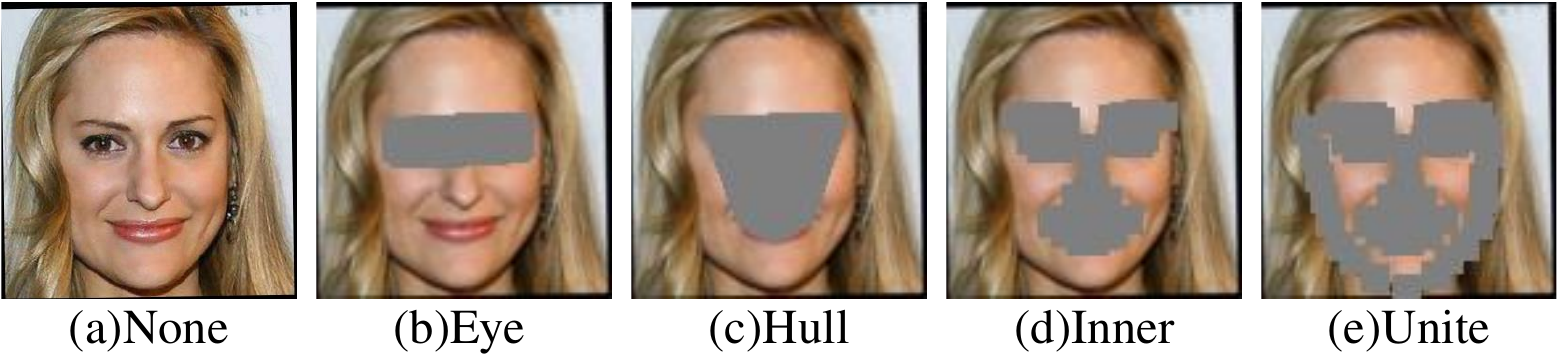}
	\vspace{-6mm}
	\caption{Visual results about different type of mask.}
	\label{fig:masktype}
\end{figure}

\begin{table}[t]
\centering
\begin{center}
\setlength{\tabcolsep}{2.7mm}{
\begin{tabular}{l|c|c|c|c}

\hline
Mask Type & FF++ &  DFD & Celeb-Df & Vox-DF \\
\hline\hline
No-mask       & 96.82 & 96.42 & 83.23 & 85.55 \\
Eye-mask      & 97.91 & 97.18 & 94.32 & 88.23\\
Hull-mask     & 96.37 & \textbf{98.07} & 96.21 & 89.95 \\
Unite-mask    & 98.02 & 96.89 & 94.45& 88.20 \\
Inner-mask    & \textbf{98.24}& 97.93 & \textbf{96.61}  &\textbf{90.61} \\
\hline
\end{tabular}}
\end{center}
\vspace{-5mm}
\caption{The performance of OuterFace by using different mask types. }
\label{mask}
\end{table}

\noindent\textbf{Masking type.}
In our default setting, we mask out some inner face regions by using the point-wise inner-mask (based on 51 landmarks) shown in \Fref{fig:masktype} (d). Here we also try other mask variants: (a) No-mask; (b)Eye-mask: only the eye part is masked; (c) Hull-mask: a landmark-based convex hull; (e) Unite-mask: masks out all the 68 landmarks.  From the results of \Tref{mask}, we have two main observations: 1) Compared to no-mask, both the point-wise based  and convex hull based masks can effectively improve DeepFake detection performance. 2) Choosing a proper mask type can help achieve better detection performance. For example, compared to inner-mask, unite-mask masks out a larger face area while its performance is a little lower.

\begin{figure}[t]
\centering
\includegraphics[width=1\columnwidth]{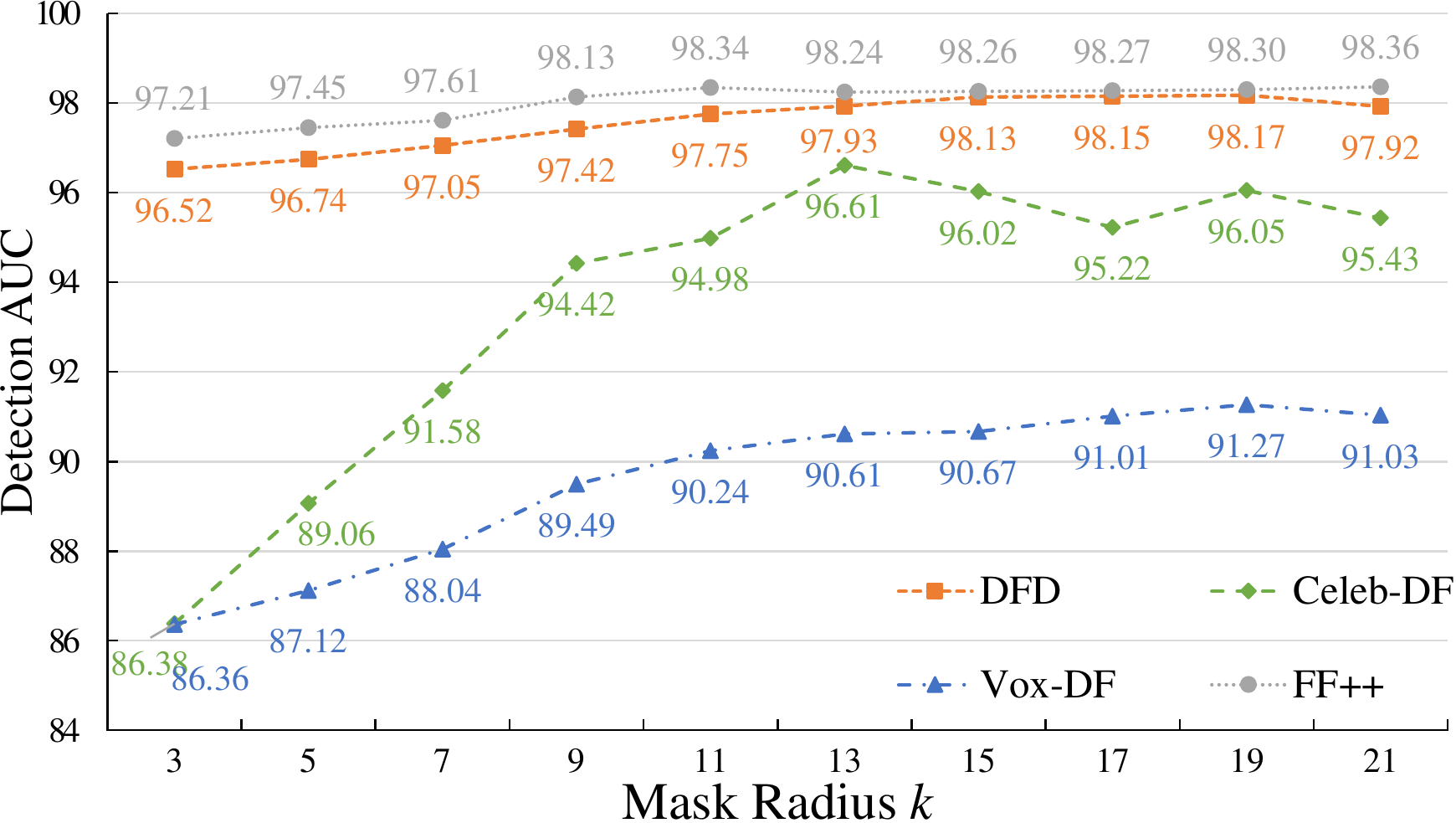} 
\vspace{-6mm}
\caption{Ablation study for the point-wise mask radius $k$}
\label{fig:radius}
\end{figure}

\begin{figure}[t]
	\centering
	\includegraphics[width=1\columnwidth]{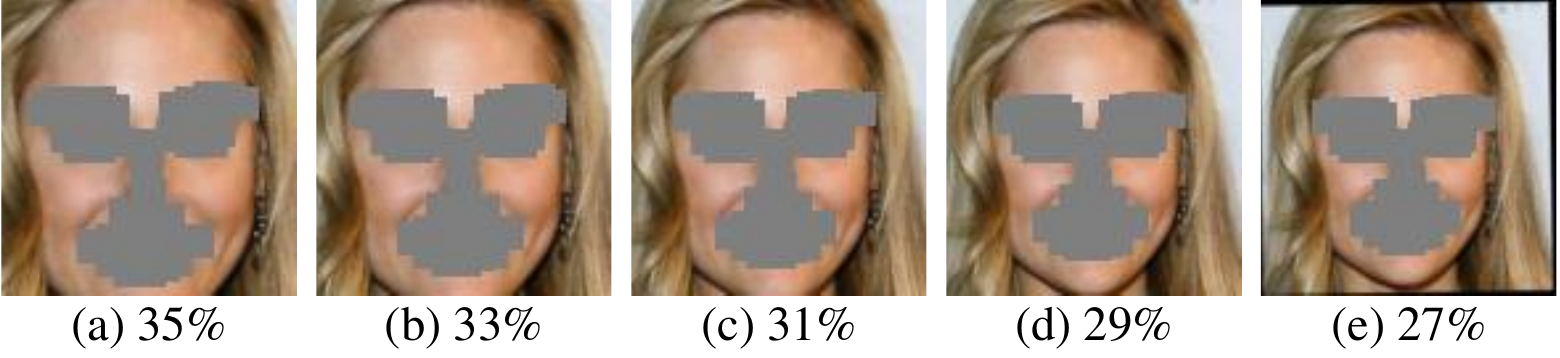}
	\vspace{-6mm}
	\caption{Example of input face area by covering different percentages of outer face region}
	\label{fig:masksize}
\end{figure}

\noindent\textbf{Point-wise mask radius $k$.}
For our default point-wise inner-mask, the mask radius $k$ controls the balance between the masked area and the visible area. If $k$ is too small, it degrades to the InnerFace variant. If $k$ is too large, the rest of the information is too little to learn a good identity embedding. In \Fref{fig:radius}, we explore the performance change of OuterFace by varying $k$ from 3 to 21. It can be seen that when $k<13$, the detection performance on all the datasets increases quickly with the increase of $k$. But when $k>13$, the detection performance on some datasets starts to slightly decrease. Therefore, we set $k=13$ as our default setting.

\begin{table}[t]
\begin{center}
\setlength{\tabcolsep}{2.8mm}{
\begin{tabular}{c|c|c|c|c}
\hline
Percentage & FF++&  DFD & Celeb-Df  & Vox-DF  \\
\hline\hline
35\% & 96.78 & 95.24 & 90.26  & 84.33 \\
33\% & 97.77 & 96.91 & 93.56 &  87.45 \\
31\% & 98.13 & 97.34 & 95.10 &  88.76\\
29\% & 98.19 & 97.36 & 95.69 & 89.20 \\
27\% & \textbf{98.24} &\textbf{97.93 }& \textbf{96.61} &  \textbf{90.61}\\
		
\hline
\end{tabular}}
\end{center}
\vspace{-5mm}
\caption{The performance of OuterFace for different percentages between the eyes distance and image width.}
\label{imsize}
\end{table}

\noindent\textbf{Face covering percentage.}
As described before, in order to cover more outer face information, we crop a large face region for training. In this ablation, we study how the area of the outer face could affect detection performance. Specifically, as shown in \Fref{fig:masksize}, we gradually increase the percentage between eye distance and image width from 27\% to 35\% with a step of 2\%. The performance difference is given in \Tref{imsize}. We find that by covering more outer face region (smaller percentage), the model will learn better.

\section{Conclusion}
In this paper, we propose a new ``identity-driven" DeepFake detection methodology. Unlike existing ``artifacts-driven'' methods, we turn DeepFake detection from an artifacts-searching problem into an identity verification problem. To facilitate research on identity-driven detection, we built a new large scale dataset “Vox-DeepFake” and present a simple and strong baseline ``OuterFace''. Even training without any fake videos, OuterFace obtains superior detection accuracy and generalization ability, which are both unachievable for existing artifact-driven methods.

\vspace{1cm}
\LARGE{\noindent\textbf{Appendix.}}
\normalsize
\section{Can OuterFace uses for face verification?}
As we introduced in Sec.\ref{sec:method} in our main paper, our OuterFace is from the idea of face verification techniques and can be viewed as an extension of traditional face verification, so it is interesting to know its performance on face verification tasks.
Here we use the classic face verification dataset Labeled Faces in the Wild (LFW) ~\cite{huang2008labeled} and calculate the accuracy follow the setting in face verification tasks. 

Results are shown in Table \ref{lfw}, we find that even we remove most of the discriminate inner face part, our OuterFace still reach a high face verification performance that 96.37\% for our default `Inner-mask' setting. Such results further proves that our OuterFace indeed learn some robust and discriminate feature from the outer part of the face that can represent the identity properly.

\section{Is hair essential for OuterFace?}
Besides the study of reference pose in our main paper, there are also some changeful attributes of reference images, such as hairstyle. However, it is hard to define the difference of hairstyles numerically, so here we evaluate the influence of hairstyles in a relatively rough way. For a given input, we use a pretrained face parsing model to calculate a hair mask and remove the hair part with it. Then we feed the masked image into our OuterFace. 

With results shown in Table \ref{hair}, we find that when we remove the hair part, the performance of our OuterFace decreases slightly. On one hand, such a decrease is reasonable as the hair-masked image is never seen during training, it would influence the performance unavoidably. On the other hand, such a small decrease further proves that our model does not rely on such changeful attributes, but learned a robust feature from the outer face part. It is also consistent with the saliency map in Figer 6 in our main paper that our OuterFace pays more attention to the unmask facial part, but not the hair. We believe if we train the model with the hair-masked image, its performance would be similar to current results while robust to the hair masking.

\begin{table}[t]
\begin{center}
\setlength{\tabcolsep}{3.7mm}{
\begin{tabular}{l|c|c|c|c}
\hline
Accuracy & Eye & Hull & Unite & Inner \\
\hline\hline
LFW      & 97.67 & 96.37 & 95.63 & 96.37 \\
\hline
\end{tabular}}
\end{center}
\caption{OuterFace performance on face verification dataset LFW.}
\label{lfw}
\end{table}

\begin{table}[t]
\begin{center}
\begin{tabular}{c|c|c|c|c}
\hline
Method &  FF++ & DFD & Celeb-Df & Vox-DF \\
\hline\hline
OutFace w/o Hair & 97.20 & 95.53 & 94.09 & 88.15\\
OuterFace        & \textbf{98.24}& \textbf{97.93} & \textbf{96.61}  & \textbf{90.61} \\
\hline
\end{tabular}
\end{center}
\caption{OuterFace performance evaluation on the hairstyle. Here w/o hair means we use a face parsing based mask to mask out the hair area.}
\label{hair}
\end{table}

\section{Implementation detail of face mask}
\begin{figure}[t]
	\centering
	\includegraphics[width=1\columnwidth]{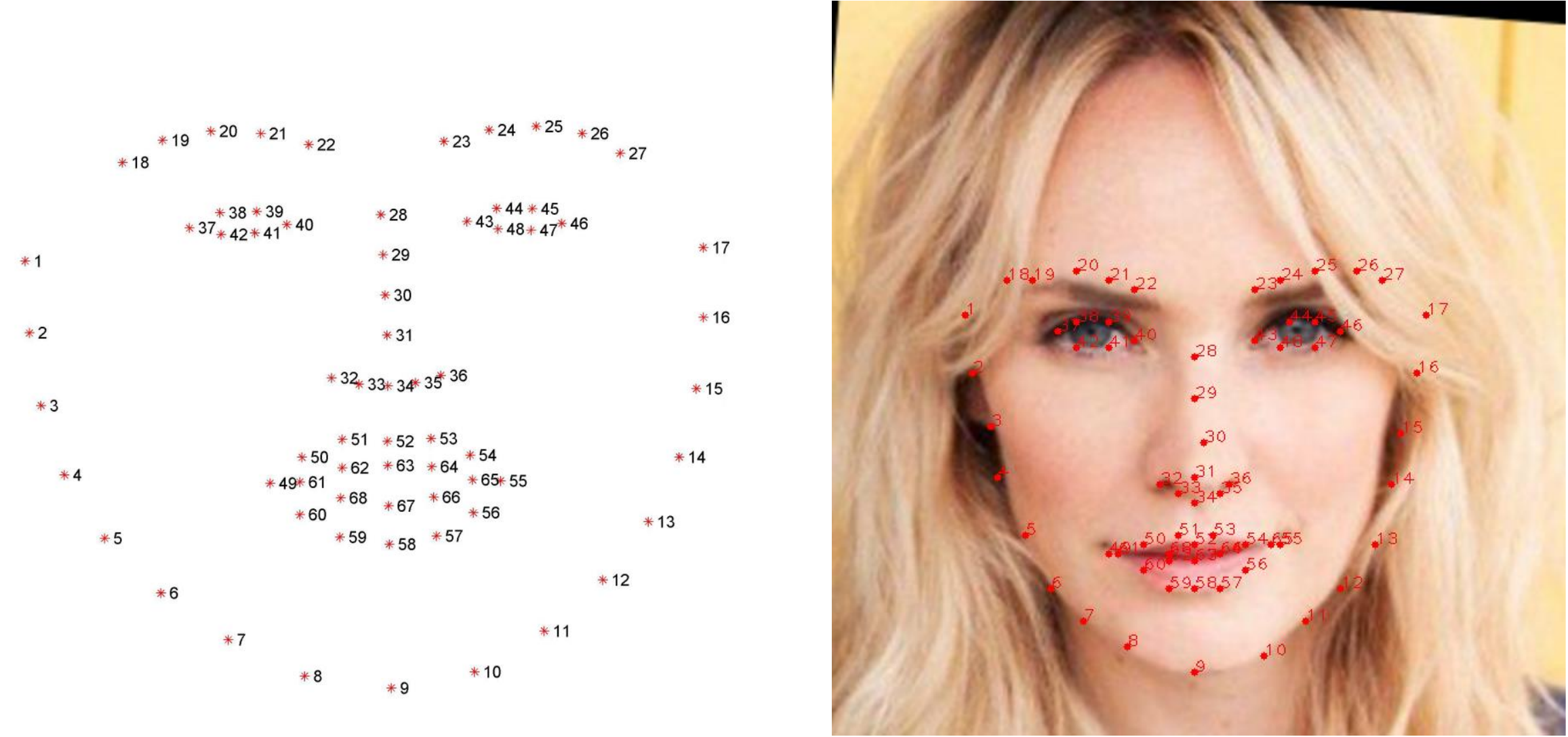} 
	\caption{Legend of the landmark for the face mask generation.}
	\label{fig:landmark}
\end{figure}

In this part, we show the detailed setting for different kinds of masks. For a clear representation, we show the index of each landmark in Figure \ref{fig:landmark}.
For the Eye-mask, we calculate the convex hull of landmarks from 37 to 48.
For the Hull-mask and Inner-mask, we use the inner face 51 landmarks start from 18 to 68. 
For the Unite-mask, we use all the 69 landmarks.

\begin{table*}[t]
\begin{center}
\begin{tabular}{c|p{1.4cm}|p{1.8cm}|p{1.4cm}|p{1.4cm}|p{1.4cm}|p{1.4cm}|p{1.4cm}|p{1.4cm}}
\hline
Method     & UADFV  & DF-TIMIT* & FF++  &  DFD & DFDC* & Celeb-DF & Deeper& Vox-DF  \\
\hline\hline
Multi-task & 57.94 & 55.30 & 72.23 & 65.21 & 53.60 & 72.28 & 65.32 & 44.60 \\
MesoInc4      & 61.82 & 62.70 & 63.41 & 59.06 & 73.20 & 42.26 & 51.41 & 54.02\\
Capsule    & 70.35 & 74.40 & 96.50 & 69.70 & 53.30 & 69.98 & 68.44 & 53.84 \\
FF-C0      & 73.53 & 54.00 & 99.26 & 89.05 & 49.90 & 48.08 & 57.76 & 47.80 \\
FF-C23     & 89.33 & 94.40 & 98.54 & 95.60 & 72.20 & 74.97 & 69.85 & 70.67 \\
FF-C40     & 83.97 & 70.50 & 93.41 & 68.80 & 69.70 & 62.04 & 72.68 & 67.09 \\
FWA        & 96.23 & 93.20 & 74.82 & 80.59 & 72.70 & 72.88 & 45.46 & 58.28 \\
DSP-FWA    & 96.29 & 99.70 & 81.90 & 90.99 & 75.50 & 78.51 & 60.00 & 45.40 \\
Face X-ray & 77.20 & -     & 98.44 & 94.14 & 71.15 & 74.76 & 72.35 & 65.08 \\
\hline
Average    & 78.51 & 75.52 & 86.56 & 79.23 & 65.69 & 66.75 & 62.58 & \textbf{56.31} \\

\hline
\end{tabular}
\end{center}
\caption{Frame-level AUC on different datasets and detection methods. Lower AUC indicates the dataset is more challenging. * indicates the results are got from other papers.}
\end{table*}

\section{Dataset evaluation}
In this part, we report the detail detection AUC of Table.1 in the main paper. Here we consider following open-source methods for the average AUC calculation: Multi-task~\cite{nguyen2019multitask}, Xception-c0,23,40~\cite{faceforensics}, FWA~\cite{li2019exposing},  DSP-FWA~\cite{li2019exposing}, Capsule ~\cite{nguyen2019use}, MesoInception4~\cite{afchar2018mesonet} (MesoInc4) and Face X-ray ~\cite{li2020face}.

{\small
\bibliographystyle{ieee_fullname}
\bibliography{egbib}
}

\end{document}